\documentclass{article}

\usepackage{arxiv}
\usepackage{graphicx}
\usepackage[utf8]{inputenc} 
\usepackage[T1]{fontenc}    
\usepackage{hyperref}       
\usepackage{url}            
\usepackage{booktabs}       
\usepackage{amsfonts}       
\usepackage{nicefrac}       
\usepackage{microtype}      
\usepackage{lipsum}

\title{Novelty search employed into the development of cancer treatment simulations}

\author{
  Michail-Antisthenis Tsompanas, Larry Bull, Andrew Adamatzky \\
  Unconventional Computing Laboratory \\
   Department of Computer Science and Creative Technologies \\
    University of the West of England\\
   Bristol, UK
     \And
     Igor Balaz\\
     Laboratory for Meteorology, Physics and Biophysics \\ 
 Faculty of Agriculture \\
University of Novi Sad\\
Novi Sad, Serbia
}

\begin{document}
\maketitle

\begin{abstract}
Conventional optimization methodologies may be hindered when the automated search is stuck into local optima because of a deceptive objective function landscape. Consequently, open ended search methodologies, such as novelty search, have been proposed to tackle this issue. Overlooking the objective, while putting pressure into discovering novel solutions may lead to better solutions in practical problems. Novelty search was employed here to optimize the simulated design of a targeted drug delivery system for tumor treatment under the PhysiCell simulator. A hybrid objective equation was used containing both the actual objective of an effective tumour treatment and the novelty measure of the possible solutions. Different weights of the two components of the hybrid equation were investigated to unveil the significance of each one.
\end{abstract}



\section{Introduction}
\label{S:1}

In optimization theory and machine learning, the idea of searching for possible solutions by putting more effort on the areas close to the optimum is well established. Nevertheless, these areas are determined based on an objective function that most definitely is riddled with local optima. It is logical that when the problem and, thus, its objective is complicated, the objective function will contain more local optima. A shortcoming of using solely an objective function is that areas in the search space, that may be stepping stones towards finding the global optimum, are neglected.

Novelty search \cite{lehman2008exploiting} was motivated by the fact that greedy search methods, which depend on a specific objective function, may suffer from deceptive evolution. Thus, the convergence to the optimum objective will in fact be hindered by this deception. Moreover, novelty search was proposed to tackle the limited advance towards higher complexity that was observed when utilizing objective-based search methods based on objective functions.

Novelty search overlooks completely the objective, while it strives towards finding something new every time. Namely, the most novel behaviour that can be derived by utilizing each solution from the search space. The fact that multiple individuals merge to a single point in the behaviour space makes the methodology computationally viable. Moreover, as multiple solutions can merge to the same point in the behaviour space, it will be expected from novelty search to continue the search towards more complex solutions. Thus, it is expected to find a good solution in the way up the complexity ladder. This enables the mitigation of the concept of open-ended search (from simulated artificial life worlds) to real problems \cite{lehman2010revising}.

This methodology managed to outperform objective driven search in several real world problems. In the study where it was first suggested, an investigation on how to design ANNs, through neuro-evolution, that navigate a robot through a maze was performed \cite{lehman2008exploiting}. Because of local optima in the objective space, namely dead-ends located close to the final target, the novelty search performed better than a greedy search. In \cite{risi2010evolving}, novelty search was implemented to a dynamic, reward-based single T-Maze problem. This kind of problem (and a couple of variations, like double T-Maze domain and a bee domain task studied in \cite{risi2010evolving}) is equipped with an essential deceptive behaviour, that the novelty search managed to handle better than the well-established objective-based evolution.

PhysiCell \cite{ghaffarizadeh2018physicell} is a multicellular, agent-based simulator that was designed to extend the BioFVM \cite{ghaffarizadeh2015biofvm} framework, to form a virtual laboratory. PhysiCell is open source and offers several sample projects, one of which is studied here. More specifically, sample project ``anti-cancer biorobots" \cite{ghaffarizadeh2018physicell} was developed as a possible tool to investigate the targeted cancer treatment, i.e. with drugs that adhere to specialized nanoparticles that would target specific molecules of the cancer cells.

The notion of PhysiCell serving as a guide to optimize the design of nanoparticle based cancer treatments \cite{PREEN20191,tsompanas2019} and discover cancer immunotherapies \cite{ozik2019learning} was previously suggested. In \cite{PREEN20191}, PhysiCell was utilized to deliver surrogate-assisted evolutionary algorithms optimising the targeted delivery of a therapeutic compound to cancerous tumour cells. In \cite{tsompanas2019} it was used, under the same application of designing a therapeutic compound delivery system, as a target simulator for a new memetic algorithm, that is inspired by the fundamental haploid-diploid lifecycle of eukaryotic organisms. Finally, in \cite{ozik2019learning}, it was combined with active learning and genetic algorithms to dynamically probe a parameter space and unveil optimal cancer regression regions of immunotherapies.

\section{Novelty search algorithm}
\label{S:2}

The implementation of novelty search is possible by utilizing any evolutionary method, while changing the objective-based fitness function with a novelty measure \cite{lehman2010efficiently}. As a result, this methodology compels the discovery of novel individuals. This new measure that will indicate how divergent each solution is compared with others in a behaviour space, should be defined based on the problem given. Choosing what the behaviour space will represent is not a priori obvious for every problem, as is the fitness function.    

The novelty measure should represent how remotely located is the behaviour of every new individual, from the rest of the so far known ones, in the behaviour space. Thus, every new individual is compared with an archive of members of the previous generations in terms of their behaviour, and not their genotypes, to determine the new individual's novelty. This archive contains individuals that were considered novel during previous generations. A well-established measure to indicate that is sparseness \cite{lehman2008exploiting}, which can be defined as the average distance of the point under study with the $k$-nearest neighbors and given in the following equation. 

\begin{equation}
\rho(x)= \frac{1}{k} \sum_{i=0}^k dist(x,\mu_i)
\label{spars}    
\end{equation}

\noindent where $\rho$ is the sparseness measure, $k$ is the number of nearest neighbors considered, $\mu_i$ is the $i$th nearest neighbor, $x$ is the individual under study and $dist$ is a function returning the distance of the two points in the behaviour space. Consequently, the individuals that are located far away from clusters of others are assigned with higher sparseness values and, thus, are considered more novel. 

Similar to a conventional fitness function space of a real problem, behaviour space can not be perfectly mapped beforehand its investigation by the evolutionary methodology. As a result, the novel individuals can be discovered only through an exploration procedure, analogous to locating the areas close to optima of the conventional objective. Moreover, novelty search has an inherent coevolutionary nature, given that the sparseness is calculated as a distance from previously discovered novel individuals throughout the evolution process. 

When a new individual has a comparably large sparseness/novelty measure value, meaning it is novel in the present generation, it is added in the aforementioned archive of novel individuals. Therefore, having this archive as a guide of where the search procedure have already sought for solutions, the methodology strives towards unexposed areas of the behaviour space, most probably containing more complex and better solutions.

\section{Methodology}
\label{S:3}

The optimization of the parameter set that determines the efficiency of worker agents, simulating the nanoparticles, in the cancer treatment simulator PhysiCell (v.1.5.1) \cite{ghaffarizadeh2018physicell} was investigated here. As population based methodologies were used, individuals were defined in a 6-D space of possible combinations of the simulator parameters. These parameters along with their ranges are: attached worker migration bias [0,1], unattached worker migration bias [0,1], worker relative adhesion [0,10], worker motility persistence time (in $mins$) [0,10], worker relative repulsion [0,10] and the cargo release $O_2$ threshold (in $mmHg$) [0,20]. All other parameters of the simulator were not modified throughout the evolutionary process and set at values same as in the initial distribution of the simulator (PhysiCell v.1.5.1 \cite{ghaffarizadeh2018physicell}), illustrated in Table \ref{tabl:1}.

\begin{table}
\centering
\caption{Unaltered parameters of PhysiCell simulator.}
\label{tabl:1}       
\begin{tabular}{|c|c|}
\hline\noalign{\smallskip}
Parameter & Value  \\
\noalign{\smallskip}\hline\noalign{\smallskip}\hline
Damage rate &  0.03333 $min^{-1}$\\ \hline
Repair rate &  0.004167 $min^{-1}$\\ \hline
Drug death rate &  0.004167 $min^{-1}$\\ \hline
Elastic coefficient &  0.05 $min^{-1}$  \\ \hline
Cargo $O_2$ relative uptake   & 0.1 $min^{-1}$ \\ \hline
Cargo apoptosis rate       & 4.065e-5 $min^{-1}$ \\  \hline
Cargo relative adhesion    & 0  \\ \hline
Cargo relative repulsion   & 5  \\ \hline
Maximum relative cell adhesion distance & 1.25 \\  \hline
Maximum elastic displacement & 50 $\mu m$  \\ \hline
Maximum attachment distance & 18 $\mu m$ \\ \hline
Minimum attachment distance & 14 $\mu m$ \\  \hline
Motility shutdown detection threshold &  0.001 \\ \hline
Attachment receptor threshold  & 0.1 \\ \hline
Worker migration speed     & 2 $\mu m / min$ \\  \hline
Worker apoptosis rate      & 0 $min^{-1}$ \\  \hline
Worker $O_2$ relative uptake  & 0.1 $min^{-1}$ \\ 
\noalign{\smallskip}\hline
\end{tabular}
\end{table}

To alleviate a part of the effect of the stochastic nature of the simulator on the results, a single tumor was used for testing every possible individual in the search space. The aforementioned tumor was produced after evolving in the simulator an initial 200$\mu m$ radius collection of cancer cells for a simulated period of 7 days. Then, for each test the fully grown tumor was loaded to the simulator (after changes in the initial source code) and the treatment was applied immediately. Namely, worker agents and cargo agents (simulating the therapeutic compound) were inserted in the simulated area. The test was finalized after 3 days from the introduction of the treatment, namely a total simulation time of 10 days from initial 200 $\mu m$ radius tumor. Nonetheless, to further minimize the effect of the stochastic procedure, the average of the outputs after 5 runs of the simulator with the same set of parameters was examined. The objective fitness of each solution was determined as the remaining cancer cells in the simulated area after the 3 days of simulated treatment.

As a reference point, the optimization of the worker agents of PhysiCell was attempted by a generic GA. {The population of the GA was of size $P=20$. The tournament method was used for parents' selection and replacement by mutated offspring with size $T=2$. 
Moreover, uniform crossover with probability $X=80\%$ was implemented and mutation rate per allele of $\mu=20\%$ with random step size of $s=[-5,5]\%$. Note that the population was evolved in generations (here for 10 generations), namely all individuals from the previous population were compared with the offspring and replaced appropriately to form the next generation.}

For the proposed methodology of novelty search, the same algorithm as in the aforementioned was used, whereas the fitness function was altered to incorporate the novelty measure. It is suggested that novelty search can be implemented in hybrid fitness functions, using both novelty measure and the objective \cite{stanley2015greatness,Mouret2011}. Using that as a motivation, we designated a hybrid fitness function as in the following:

\begin{equation}
fitness= \frac{rcc}{rcc_{thr}} - \frac{sparseness}{s_{thr}}
\label{fitns}    
\end{equation}

\noindent where $rcc$ is the remaining number of cancer cells after the 3 days of the cancer treatment and $sparseness$ the average distance of the new individual's behaviour from the 5 nearest neighbors in the behaviour space (as defined in Eq. \ref{spars}). Moreover, $rcc_{thr}$ and $s_{thr}$ are parameters used to normalize the values of the remaining number of cancer cells and sparseness, defined in the following experiments as 1400 for the first parameter and in the range of 200 to 1000 with intervals of 200 for the second one.

The output of each solution in the behaviour space was defined as the center of gravity of the ensemble of worker agents at their final position after 3 days of simulated treatment. More specifically, the placement of the collection of nanoparticles in the simulated area. This behaviour is easily calculated by the average of the coordinates of all the worker agents. Consequently, as in previous works of novelty search \cite{lehman2008exploiting, Gomes2013}, the topology of the result was taken into account, which is ignoring the actual objective.

\section{Results}
\label{S:4}

\begin{figure}[!t]
\centering
\includegraphics[width=0.5\linewidth]{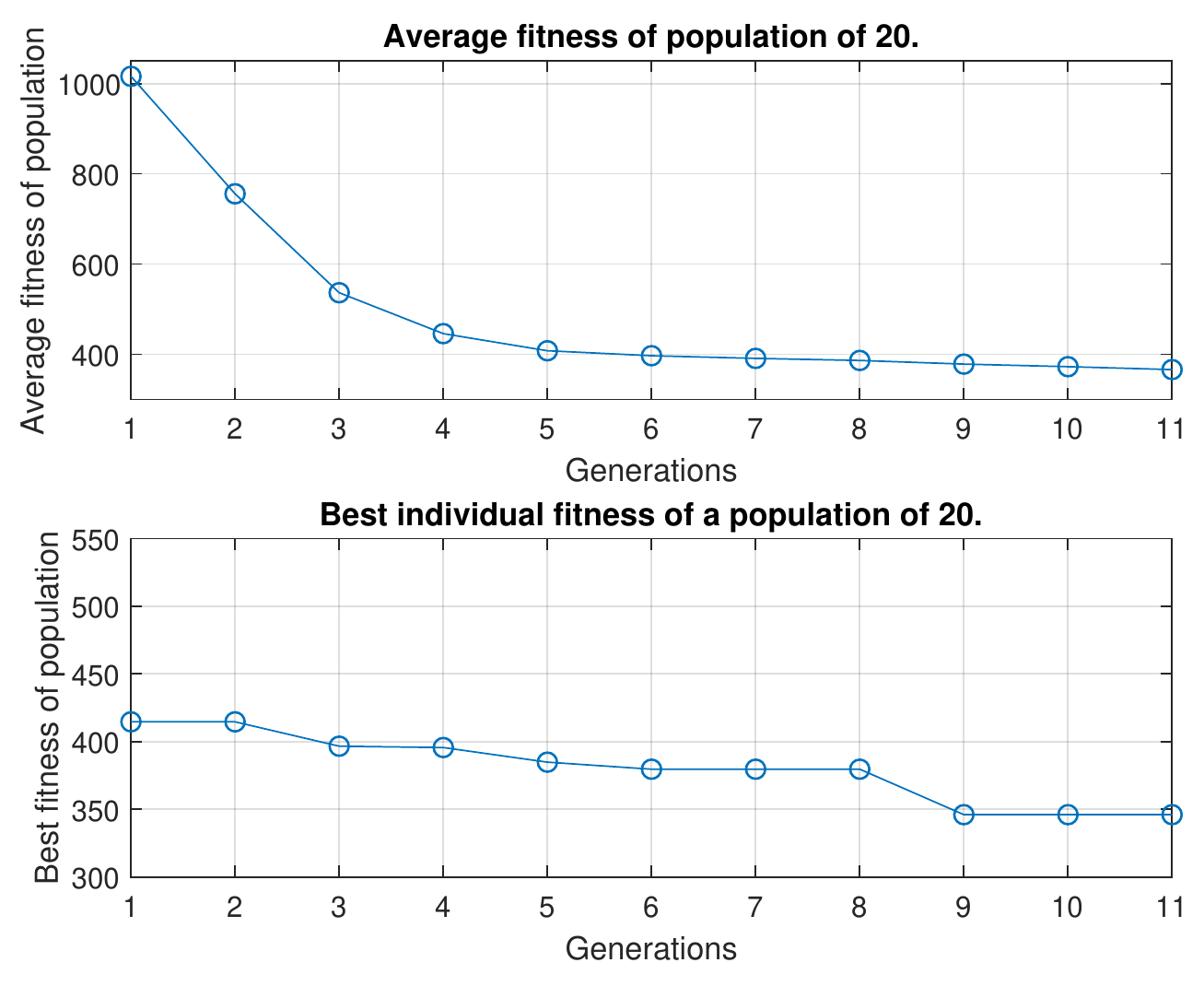}
\caption{Average and best actual fitness of individuals in each generation for the simple GA.}
\label{c0}
\end{figure}

\begin{figure}[!t]
\centering
\includegraphics[width=0.5\linewidth]{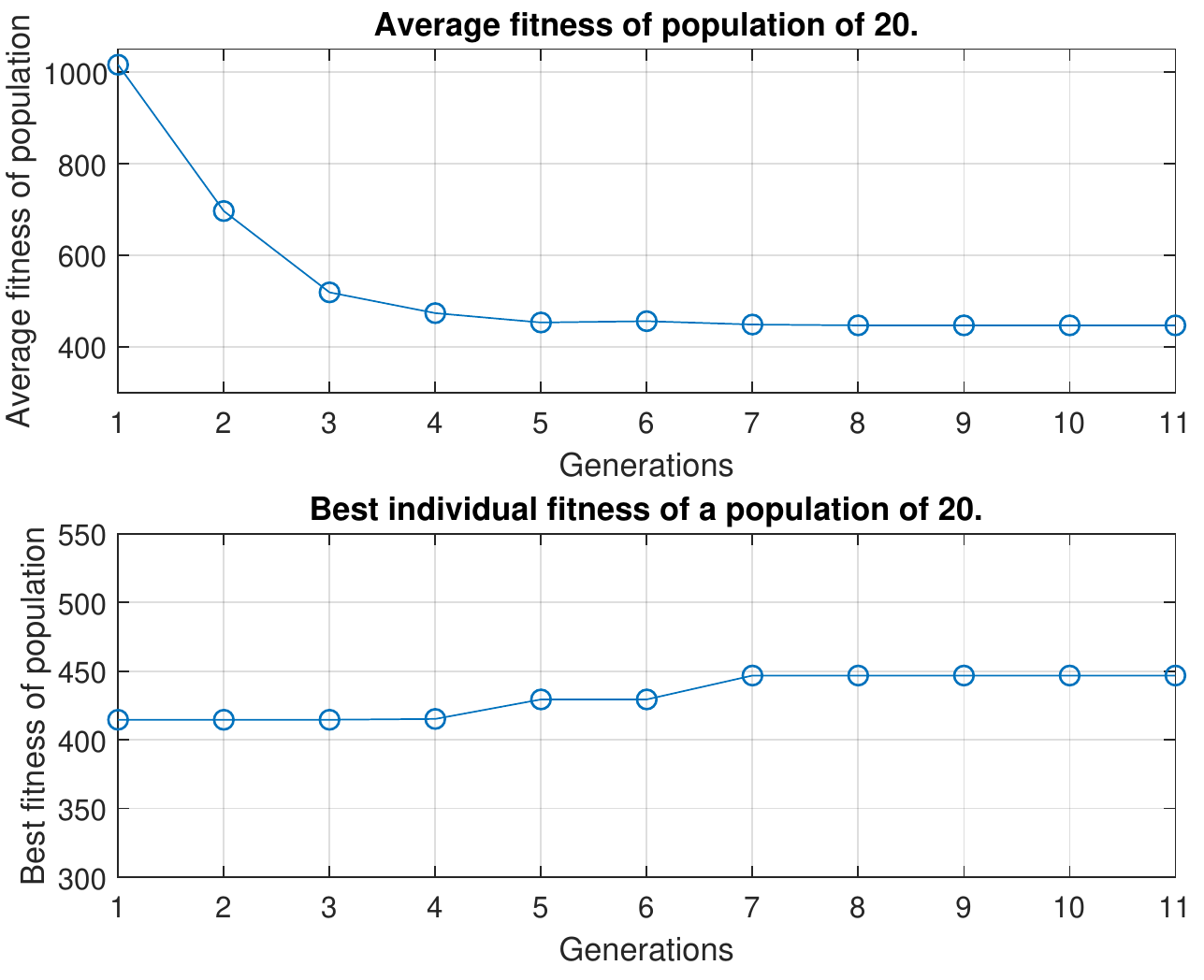}
\caption{Average and best actual fitness of individuals in each generation for the hybrid fitness function with $s_{thr}=200$.}
\label{c200}
\end{figure}

\begin{figure}[!t]
\centering
\includegraphics[width=0.5\linewidth]{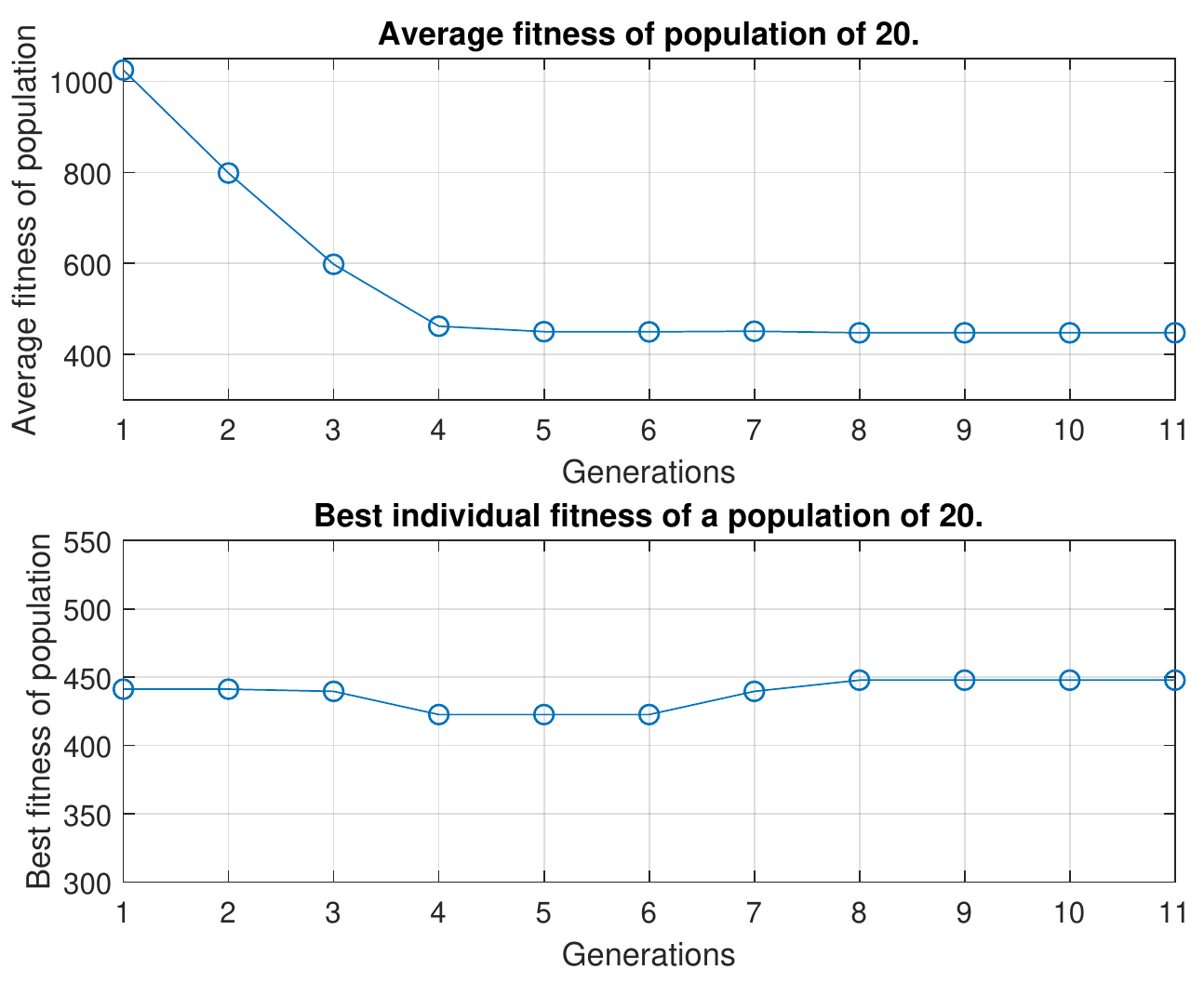}
\caption{Average and best actual fitness of individuals in each generation for the hybrid fitness function with $s_{thr}=400$.}
\label{c400}
\end{figure}

\begin{figure}[!t]
\centering
\includegraphics[width=0.5\linewidth]{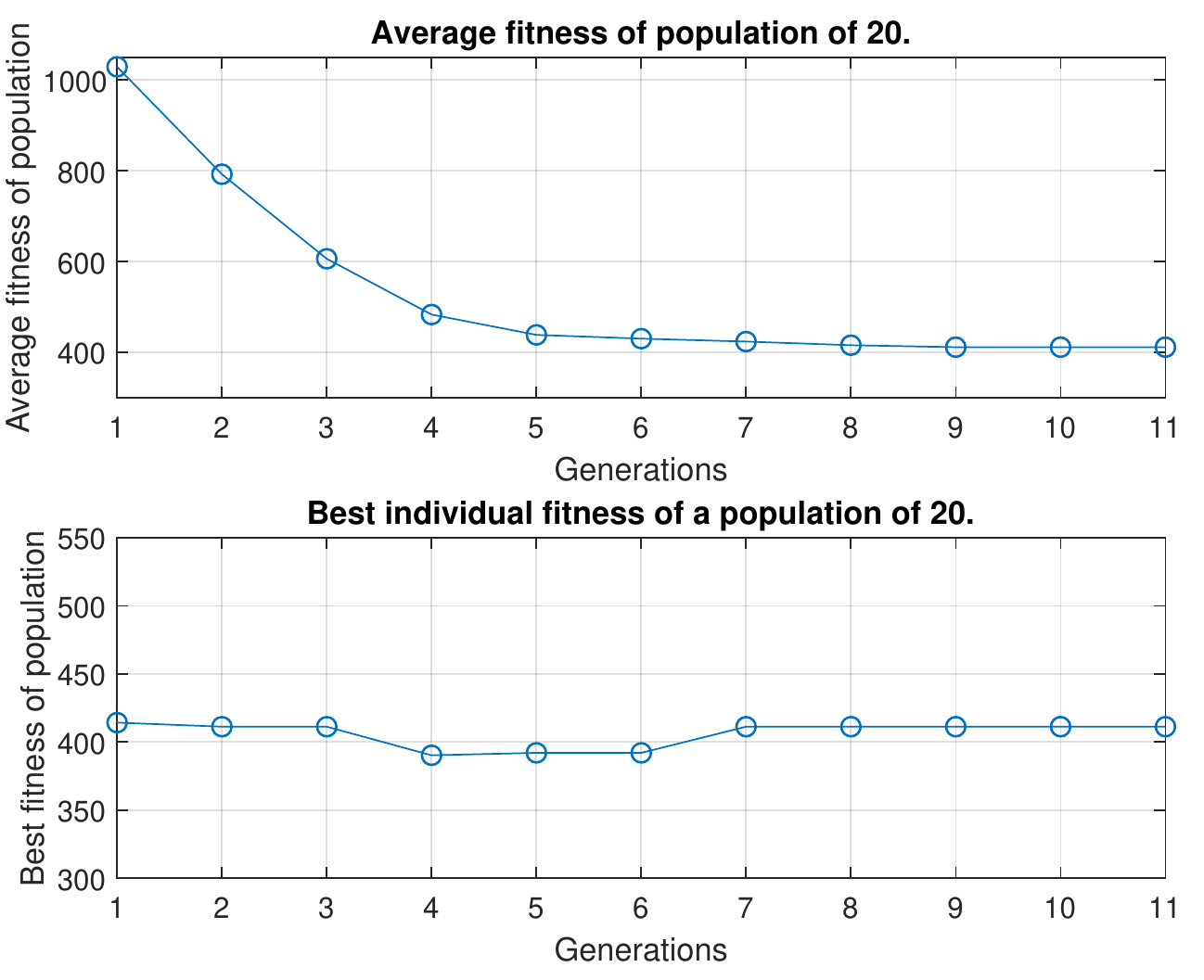}
\caption{Average and best actual fitness of individuals in each generation for the hybrid fitness function with $s_{thr}=600$.}
\label{c600}
\end{figure}

\begin{figure}[!t]
\centering
\includegraphics[width=0.5\linewidth]{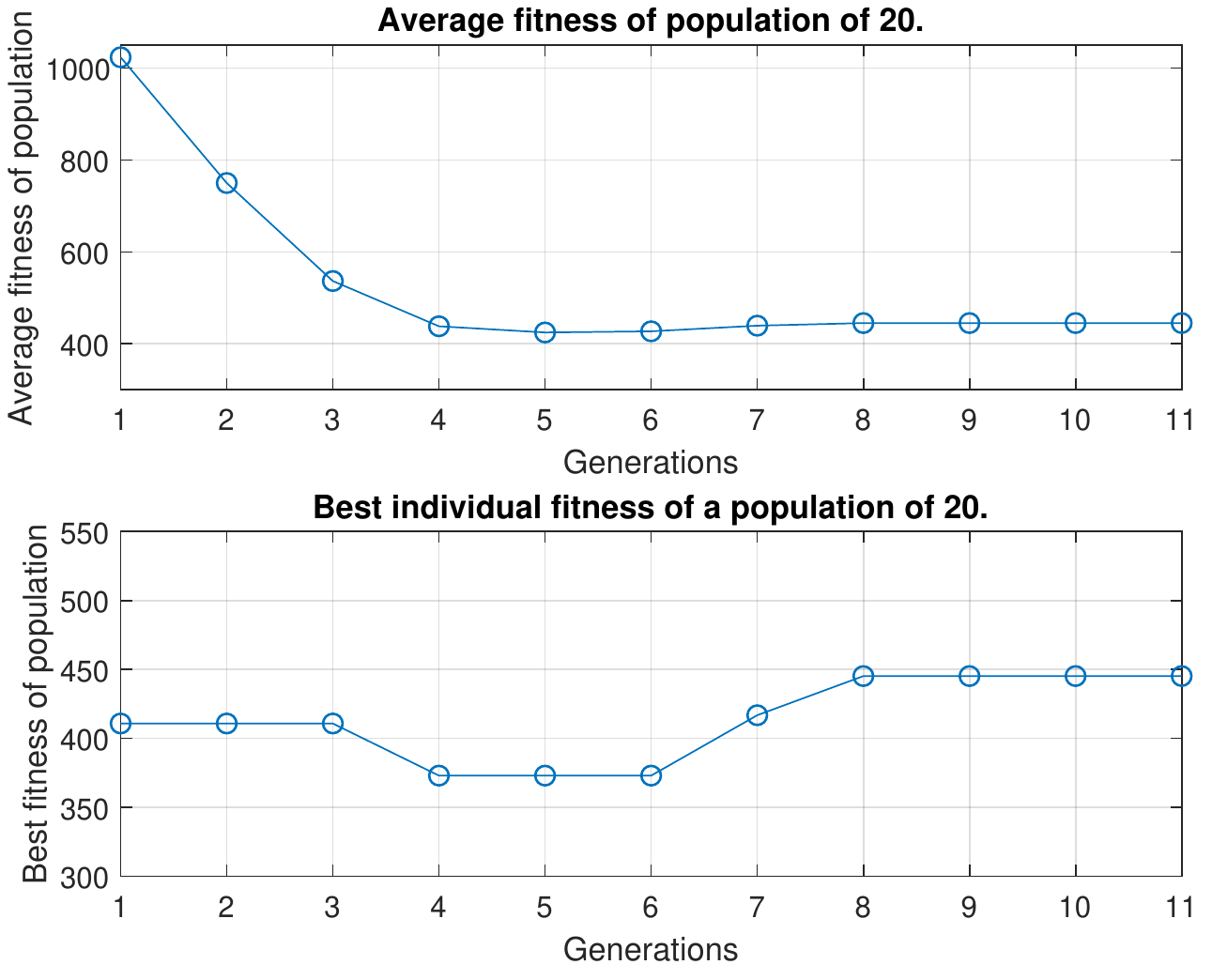}
\caption{Average and best actual fitness of individuals in each generation for the hybrid fitness function with $s_{thr}=800$.}
\label{c800}
\end{figure}

\begin{figure}[!t]
\centering
\includegraphics[width=0.5\linewidth]{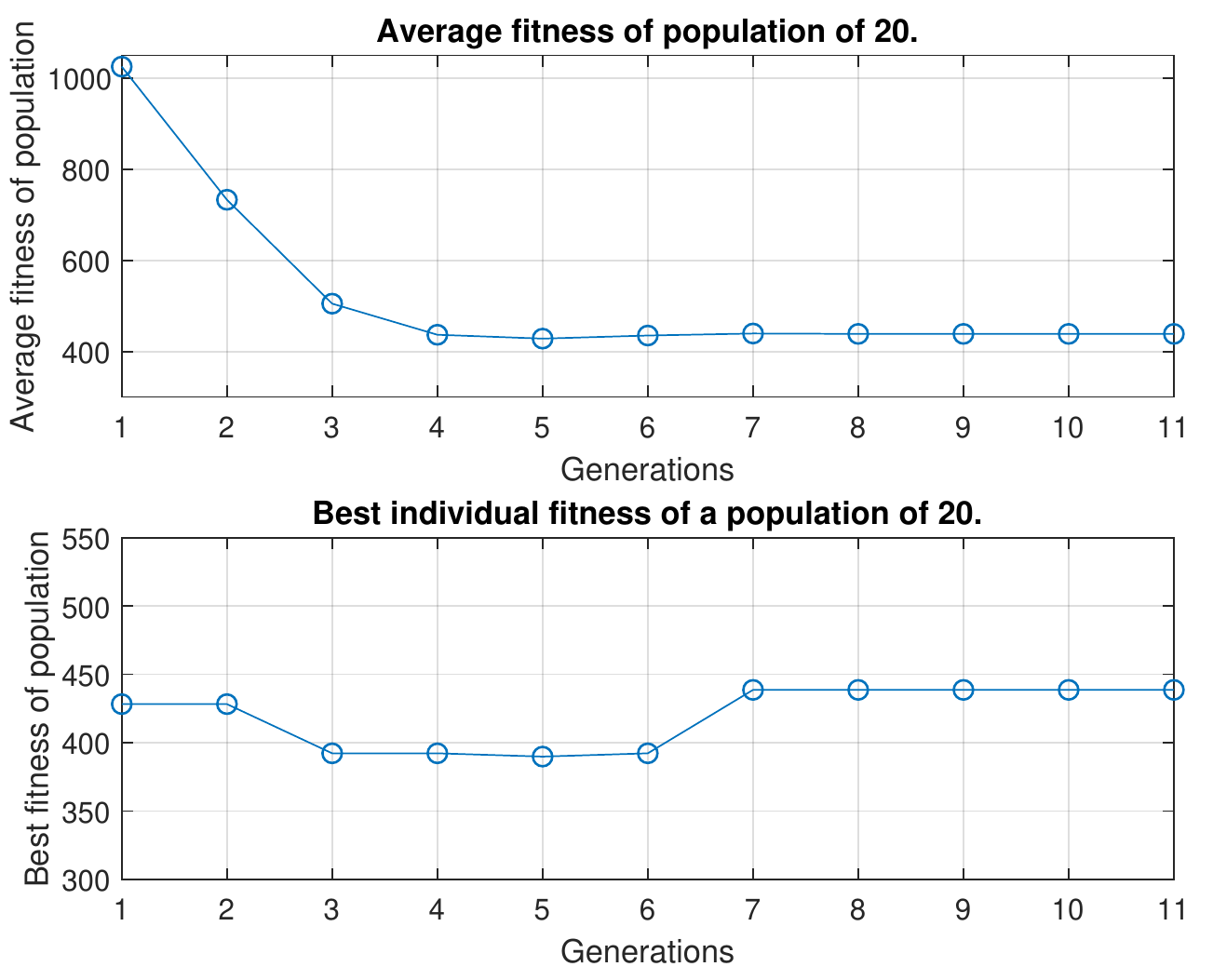}
\caption{Average and best actual fitness of individuals in each generation for the hybrid fitness function with $s_{thr}=1000$.}
\label{c1000}
\end{figure}

To make the comparison between different algorithms meaningful the initial population for every case is composed by the same individuals. Three different sets (of $P=20$ individuals) of initial populations were tested. The outputs of using a generic GA and the hybrid fitness function (as described in Eq. \ref{fitns}) with different normalization parameter $s_{thr}$ are depicted in Figs. \ref{c0} - \ref{c1000}. These figures illustrate the average actual fitness of the population for every generation and the actual fitness of the best individual found in each generation. By the term actual fitness, we define the number of remaining cancer cells in the simulated area, not to be confused with the hybrid fitness function used in novelty search method and given in Eq. \ref{fitns}.

From the results in Figs. \ref{c0} - \ref{c1000} it is established that while the simple GA provides a better (or at least the same) fitness for every generation, the novelty search method presents a more erratic behaviour. Namely, with $s_{thr}=200$ it does not manage to find a better solution than the initial randomly generated one, on the contrary it searches the solution landscape without any profound advance in fitness. However, for higher values of the parameter $s_{thr}$  (meaning smaller significance of the novelty measure compared with the actual fitness), the searching method manages to optimize the solution at least briefly in the extend of the 10 generations. More specifically, as depicted in Figs. \ref{c400} - \ref{c1000} there is a decline in the amount of remaining cancer cells (actual fitness) for the up to the 6th generation, but then the novelty measure seems to be putting more pressure into finding more novel solutions than remaining the fittest in the population. The decline in the actual fitness is more profound for the middle values in the range of the $s_{thr}$ parameter, specifically for $s_{thr}=800$.

\begin{figure}[!t]
\centering
\includegraphics[width=0.5\linewidth]{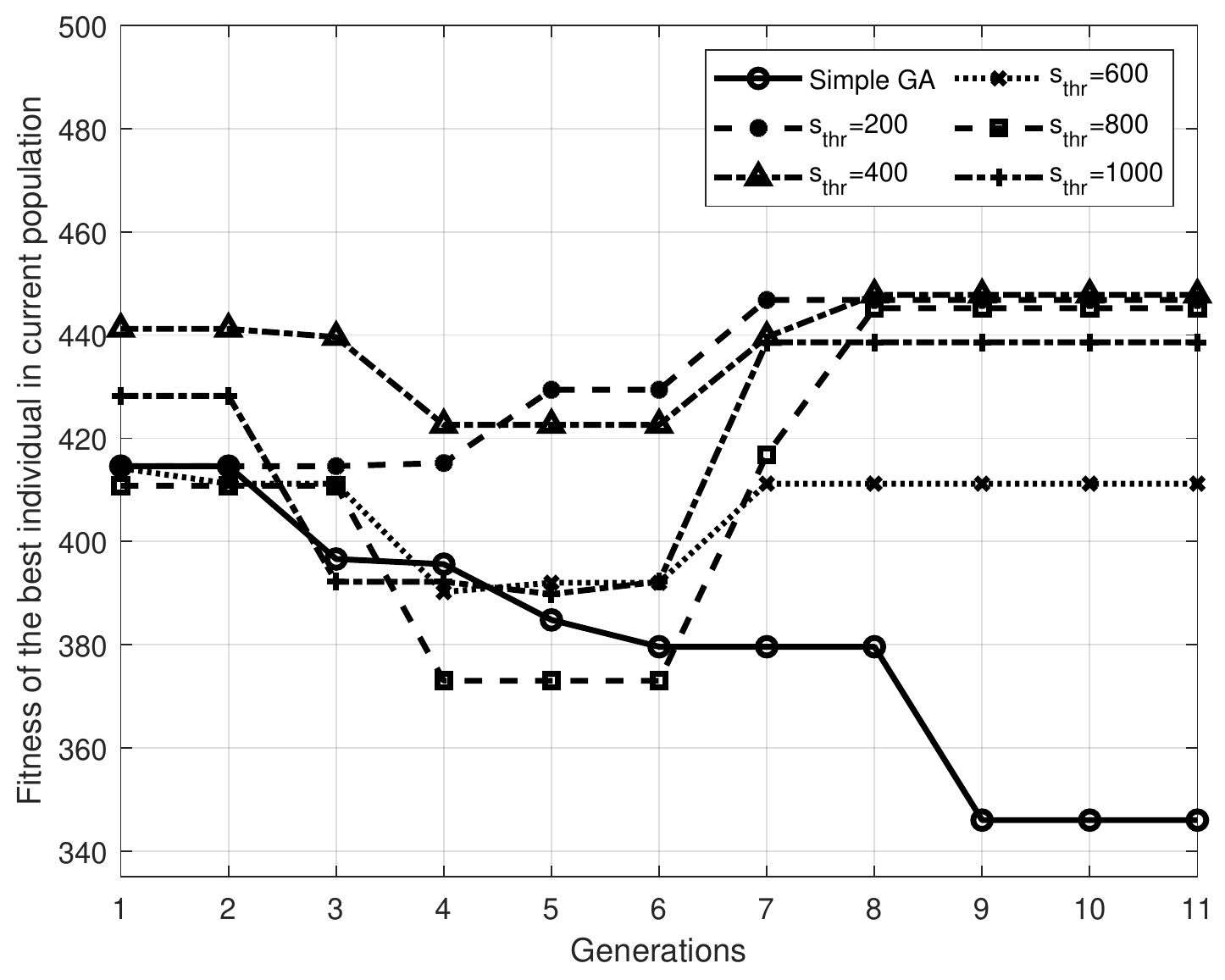}
\caption{Cumulative results of first run. Best actual fitness of individuals in each generation for all $s_{thr}$ parameters compared with the simple GA.}
\label{cComp1}
\end{figure}

\begin{figure}[!t]
\centering
\includegraphics[width=0.5\linewidth]{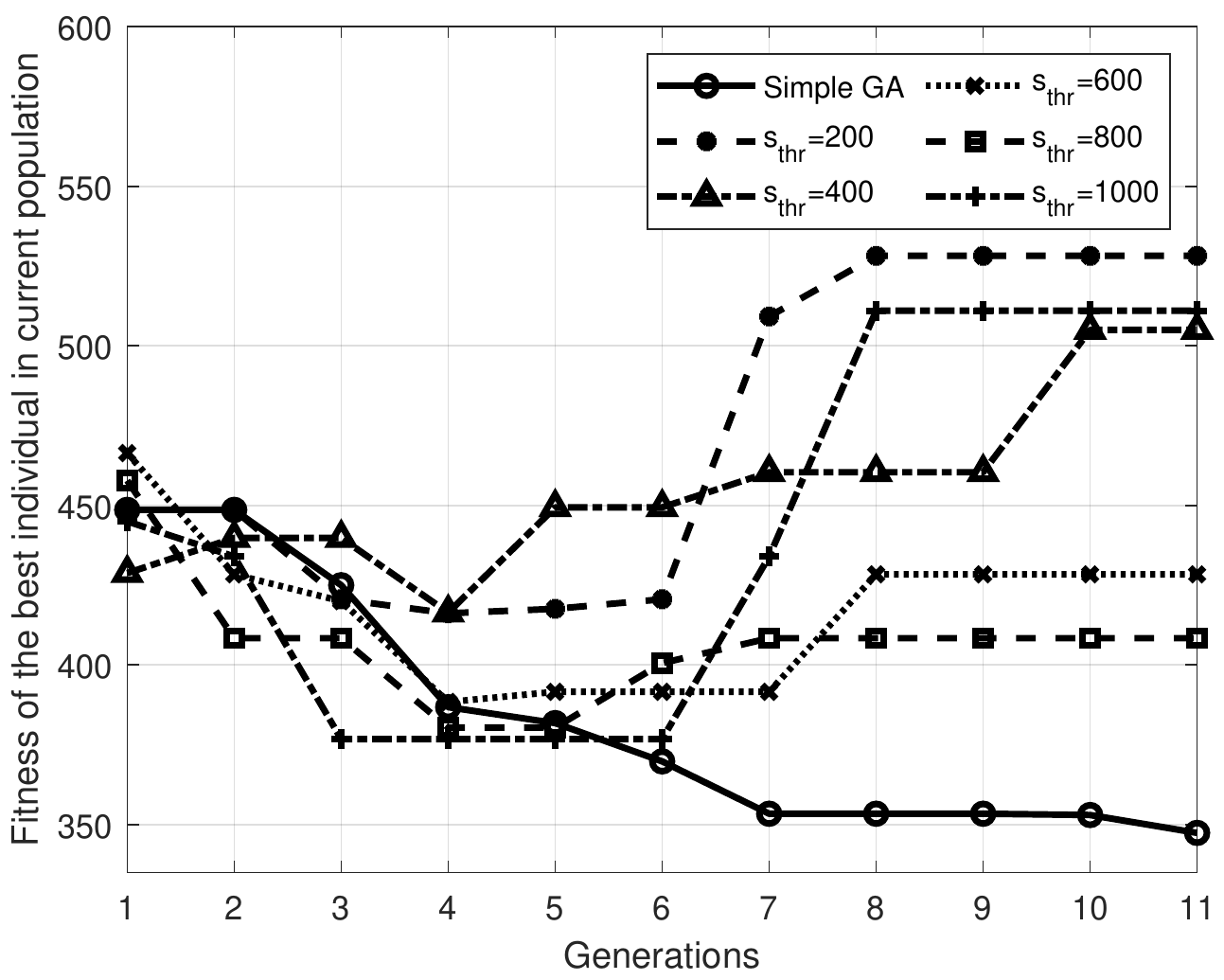}
\caption{Cumulative results of second run. Best actual fitness of individuals in each generation for all $s_{thr}$ parameters compared with the simple GA.}
\label{aComp1}
\end{figure}

\begin{figure}[!t]
\centering
\includegraphics[width=0.5\linewidth]{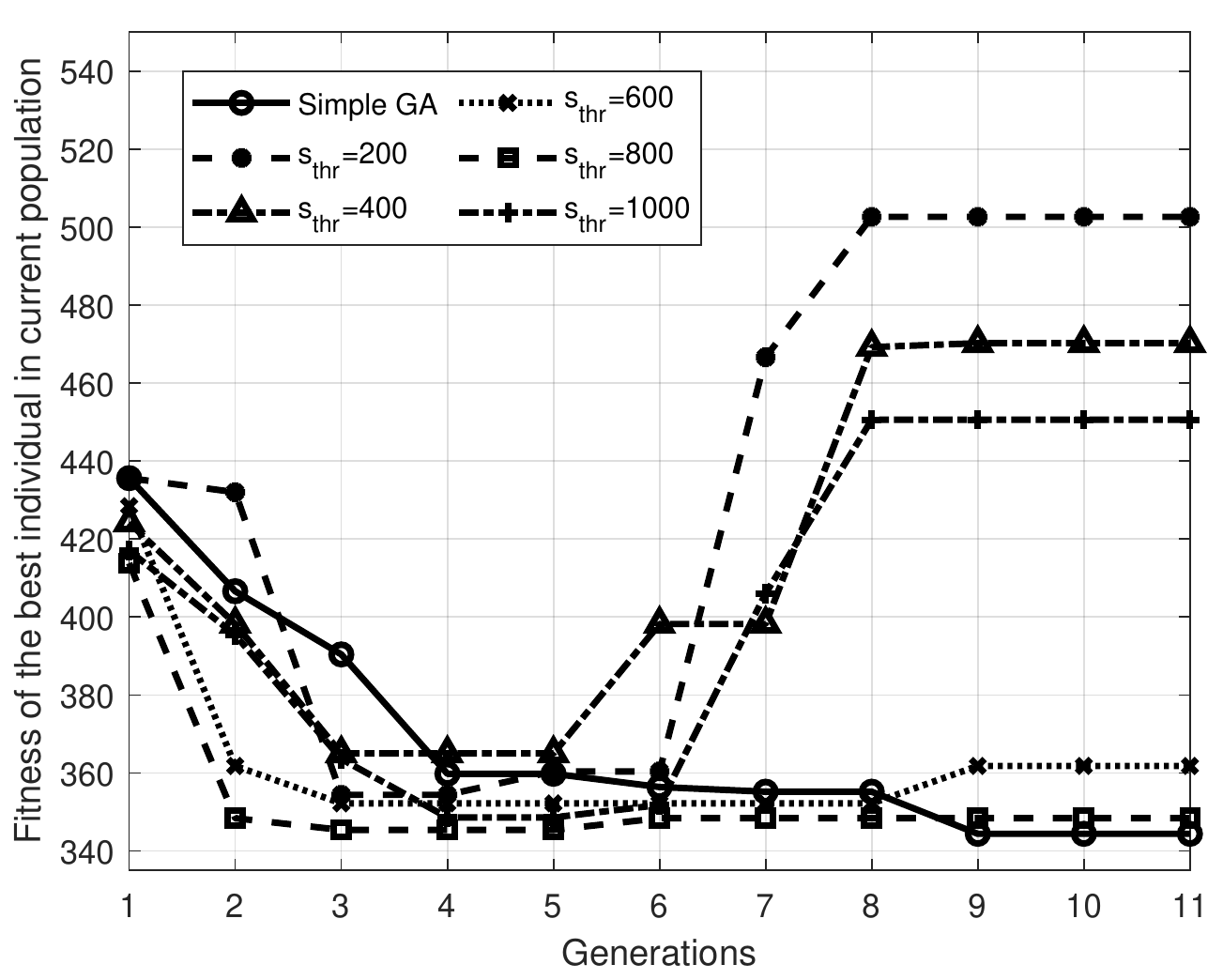}
\caption{Cumulative results of third run. Best actual fitness of individuals in each generation for all $s_{thr}$ parameters compared with the simple GA.}
\label{bComp1}
\end{figure}

To better compare the results of the search method with different $s_{thr}$ parameters and the simple GA, Figs. \ref{cComp1} - \ref{bComp1} are provided. Each figure is containing the results of every run, namely the use of different methods in the same initial population of $P=20$ individuals.

Throughout all the different runs, it can be observed that novelty search yields more erratic outputs when studying the actual fitness. An outcome that is expected, given the fact that the hybrid fitness function used in this search method contains the novelty measure that completely ignores the actual fitness of the solutions. Nevertheless, it can be noticed that while in the final results (after 10 generations of artificial evolution) simple GA is providing better solutions, in most of the initial generations, novelty search is providing better solutions. In particular for $s_{thr}$ parameters higher than 400.

\begin{figure}[!t]
\centering
\includegraphics[width=0.5\linewidth]{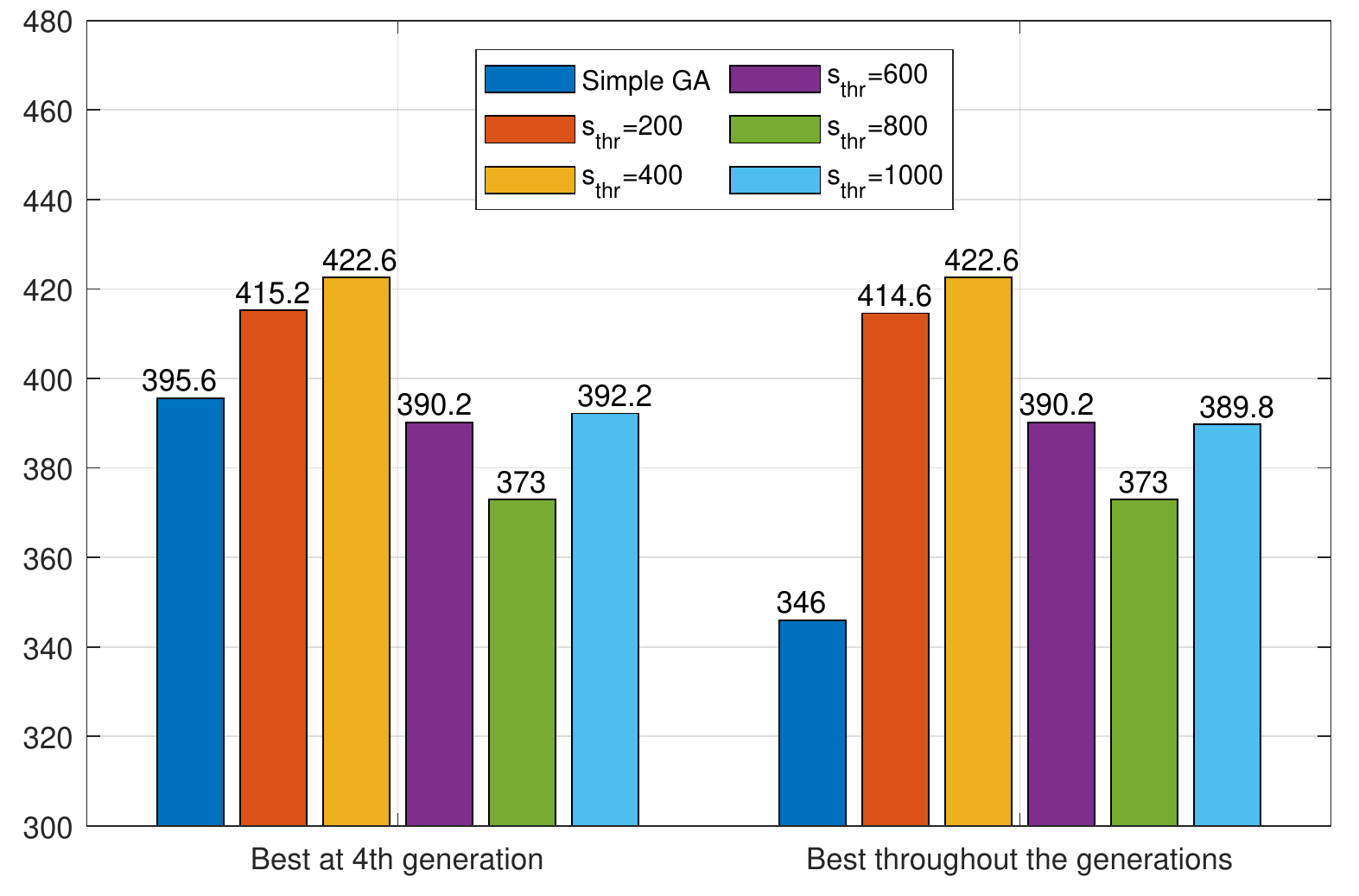}
\caption{Best actual fitness of individuals in 4th generation and throughout all generations.}
\label{cBars1}
\end{figure}

This fact is better illustrated in Fig. \ref{cBars1}. Here the best individual discovered until the 4th generation and throughout all the generations is presented for the simple GA and the different cases of the hybrid novelty search for the first run. Despite the fact that the simple GA seems to outperform the novelty search throughout the 10 generations, it seems that the novelty search with $s_{thr} \geq  600$ outperforms the simple GA for up to the 4th generation.

\begin{figure}[!t]
\centering
\includegraphics[width=0.5\linewidth]{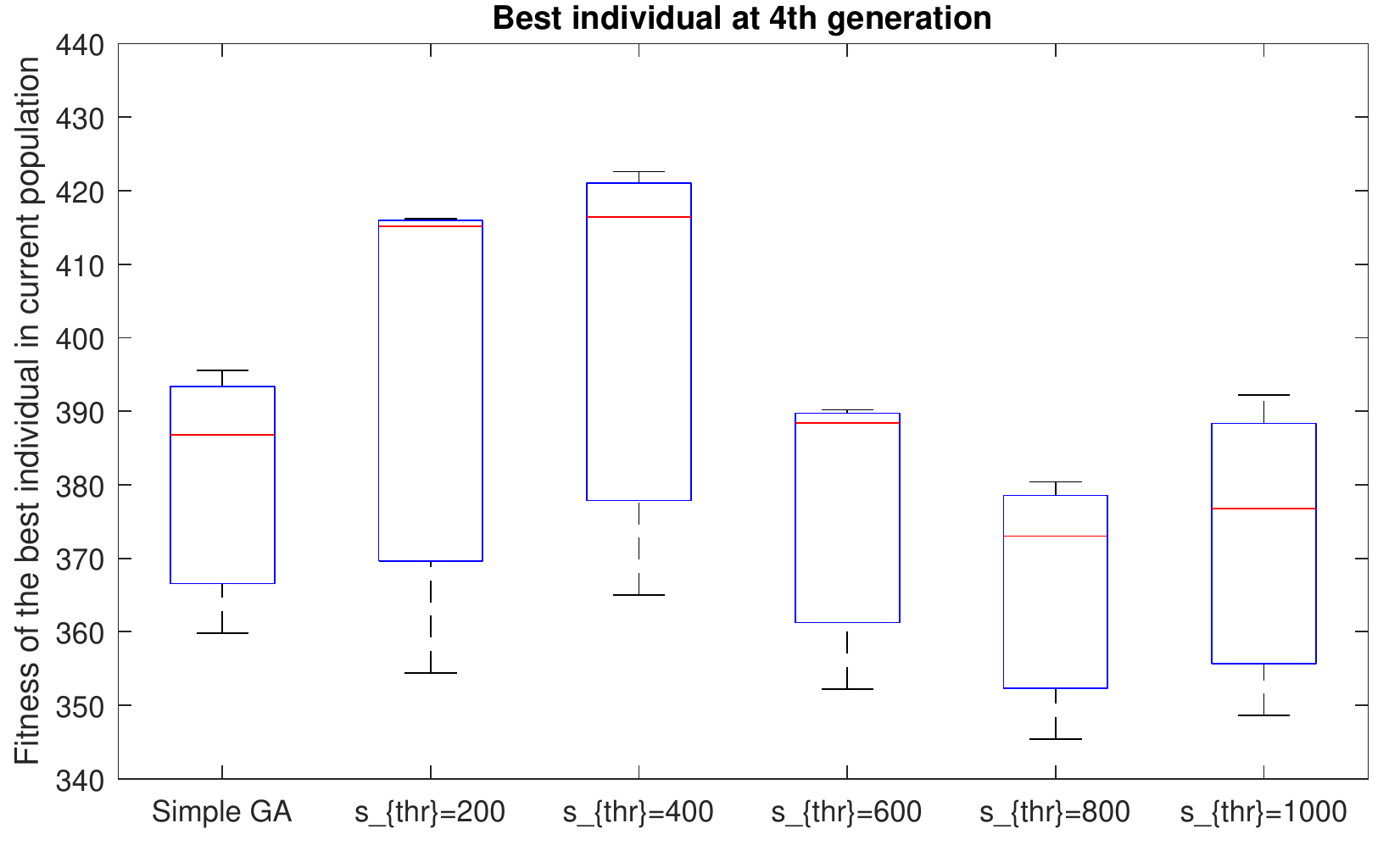}
\caption{Boxplot of the best actual fitness of individuals in 4th generation for all runs.}
\label{box4}
\end{figure}

\begin{figure}[!t]
\centering
\includegraphics[width=0.5\linewidth]{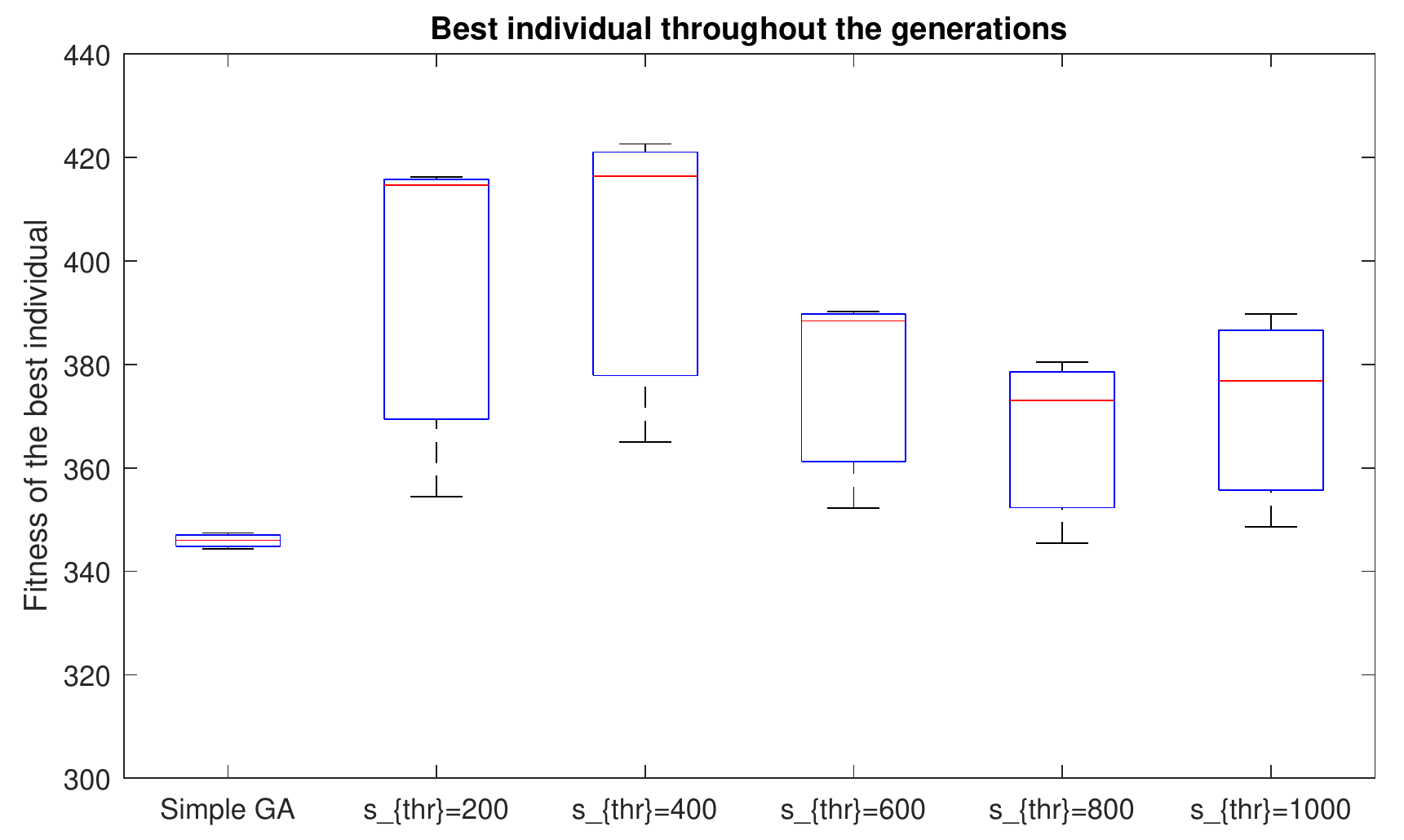}
\caption{Boxplot of the best actual fitness of individuals in all generations for all runs.}
\label{boxAll}
\end{figure}

The same finding stands for all three runs (different initialization of the comparison test). This can be realized by Figs. \ref{box4} and \ref{boxAll} rendering the boxplots of the best individual in terms of actual fitness up until the 4th generation and throughout the length of the all the generations. 

\section{Conclusion}
\label{S:5}

Novelty search is motivated by the need to overcome the problems of deception and local optima inherent in objective optimization. Ignoring the objective completely or using hybrid fitness functions including a novelty measure, may often benefit the search of a better solution. In this study, this methodology was employed to optimize the design of targeted drug delivery systems, aiming cancerous tumours. The solutions were evaluated by PhysiCell simulator, namely by its sample project ''anti-cancer biorobots''. The association of the fitness function with a novelty measure rather than only the objective proved to lead to more efficient solutions faster in the initial steps of artificial evolution. Moreover, an analysis of the significance of the novelty measure was performed by running optimization processes with different weights on the novelty measure. The medium and high values in the range studied proved to be more effective. 

Nonetheless, novelty search has some limitations. Given the fact that it ignores the objective, there is no pressure towards further optimization once a good but not ideal solution is found. An optimized solution may be produced by novelty search only if an individual can appear novel while demonstrating this optimized performance. As illustrated in the results provided, a simple GA was able to outperform the hybrid novelty search in the course of 10 evolution steps. A possible solution to this limitation, is to take the most promising results from novelty search and further optimize them based on an objective function. Thus, following this procedure will take advantage of the strengths of both approaches. Novelty search successfully locates the approximate solutions, while objective optimization further investigates the close area around approximate solutions. 

On the other hand, novelty search can be applied in the case of a traditional evolutionary algorithm reaches convergence, to inject the population with new diversified individuals. These prospects of combined novelty and objective based procedures can serve as aspects of future work. Finally, the conclusions driven from this study will be applied on ongoing research \cite{PREEN20191,tsompanas2019} towards a more wide applicability platform that will design, develop and evaluate DDSs aiming cancer tumours.

\section{Acknowledgement}

This work was supported by the European Research Council under the European Union's  under grant agreement No. 800983.

\bibliographystyle{unsrt}

\end{document}